\documentclass[runningheads]{llncs}
\usepackage[T1]{fontenc}
\usepackage[pagebackref=true,breaklinks=true,colorlinks,bookmarks=false]{hyperref}
\usepackage{comment}
\usepackage{bm}
\usepackage{booktabs}
\usepackage{bold-extra}
\usepackage{xspace}
\usepackage[misc]{ifsym}

\usepackage[table,xcdraw]{xcolor}
\usepackage{dsfont}
\usepackage{colortbl}
\usepackage[frozencache]{minted}
\usepackage{textcomp}
\usepackage{multirow}
\usepackage{algorithm}
\usepackage{algorithmic}
\usepackage{amsmath}
\usepackage{makecell}
\usepackage{amssymb}
\usepackage[figuresright]{rotating}
\usepackage{xspace}
\newcommand{\eg}{\emph{e.g.}\xspace}

\begin{document}
%
\title{Mining Negative Temporal Contexts\\For False Positive Suppression\\In Real-Time Ultrasound Lesion Detection}
\titlerunning{Mining Negative Temporal Contexts To Suppress FPs}
%
\author{Haojun Yu\inst{1} \and
Youcheng Li\inst{1}\and
QuanLin Wu\inst{2,5} \and
Ziwei Zhao\inst{2} \and\\
Dengbo Chen\inst{4} \and
Dong Wang\inst{1} \and
Liwei Wang\inst{1,3(\textrm{\Letter})}
}
%
\authorrunning{H. Yu et al.}
%
\institute{National Key Laboratory of General Artificial Intelligence, School of Intelligence Science and Technology, Peking University, Beijing, China\\
\email{\{haojunyu, wanglw\}@pku.edu.cn}\and
Center of Data Science, Peking University, Beijing, China\and
Center for Machine Learning Research,  Peking University, Beijing, China\and
Yizhun Medical AI Co., Ltd, Beijing, China\and
Pazhou Laboratory (Huangpu), Guangdong, China
}
\maketitle              
\begin{abstract}

During ultrasonic scanning processes, real-time lesion detection can assist radiologists in accurate cancer diagnosis. However, this essential task remains challenging and underexplored. General-purpose real-time object detection models can mistakenly report obvious false positives (FPs) when applied to ultrasound videos, potentially misleading junior radiologists. One key issue is their failure to utilize negative symptoms in previous frames, denoted as \emph{negative temporal contexts} (NTC)~\cite{spak2017bi}. To address this issue, we propose to extract contexts from previous frames, including NTC, with the guidance of inverse optical flow. By aggregating extracted contexts, we endow the model with the ability to suppress FPs by leveraging NTC. We call the resulting model \emph{UltraDet}. The proposed UltraDet demonstrates significant improvement over previous state-of-the-arts and achieves real-time inference speed. We release the code, checkpoints, and high-quality labels of the CVA-BUS dataset~\cite{lin2022new} in \url
{https://github.com/HaojunYu1998/UltraDet}.

\end{abstract}
\section{Introduction}
\begin{figure}[t]
    \centering
    \includegraphics[width=0.9\textwidth]{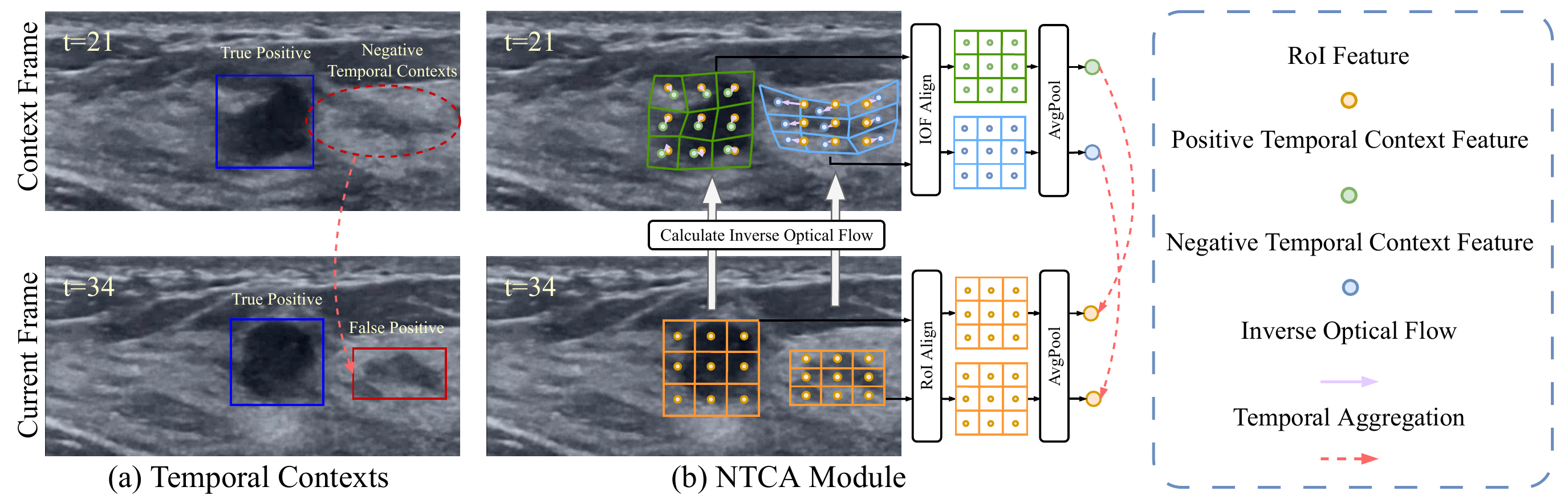}
    \caption{Illustration of Negative Temporal Context Aggregation (NTCA) module. (a) Our motivation: mining negative temporal contexts for FP suppression. (b) The NTCA module leverages temporal contexts to suppress the FP.}
    \label{fig:teaser}
\end{figure}

Ultrasound is a widely-used imaging modality for clinical cancer screening. Deep Learning has recently emerged as a promising approach for ultrasound lesion detection. While previous works focused on lesion detection in still images~\cite{yap2017automated} and offline videos~\cite{movahedi2020automated,lin2022new,wang2022key}, this paper explores real-time ultrasound video lesion detection. Real-time lesion prompts can assist radiologists during scanning, thus being more helpful to improve the accuracy of diagnosis. This task requires the model to infer faster than 30 frames per second (FPS)~\cite{vaze2020low} and only previous frames are available for current frame processing.

Previous general-purpose detectors~\cite{deng2019relation,chen2020memory} report simple and obvious FPs when applied to ultrasound videos, \eg the red box in Figure~\ref{fig:teaser}(a). These FPs, attributable to non-lesion anatomies, can mislead junior readers. These anatomies appear like lesions in certain frames, but typically show negative symptoms in adjacent frames when scanned from different positions. So experienced radiologists will refer to corresponding regions in previous frames, denoted as \emph{temporal contexts} (TC), to help restrain FPs. If TC of a lesion-like region exhibit negative symptoms, denoted as \emph{negative temporal contexts} (NTC), radiologists are less likely to report it as a lesion~\cite{spak2017bi}. Although important, the utilization of NTC remains unexplored. In natural videos, as transitions from non-objects to objects are implausible, previous works~\cite{deng2019relation,chen2020memory,wang2022ptseformer} only consider inter-object relationships. As shown in Section~\ref{subsec:main_result}, the inability to utilize NTC is a key issue leading to the FPs reported by general-purpose detectors.

To address this issue, we propose a novel \emph{UltraDet} model to leverage NTC. For each Region of Interest (RoI) $\mathcal{R}$ proposed by a basic detector, we extract temporal contexts from previous frames. To compensate for inter-frame motion, we generate deformed grids by applying inverse optical flow to the original regular RoI grids, illustrated in Figure~\ref{fig:teaser}. Then we extract the RoI features from the deformed grids in previous frames and aggregate them into $\mathcal{R}$. We call the overall process \emph{Negative Temporal Context Aggregation} (NTCA). The NTCA module leverages RoI-level NTC which are crucial for radiologists but ignored in previous works, thereby effectively improving the detection performance in a reliable and interpretable way. We plug the NTCA module into a basic real-time detector to form \emph{UltraDet}. Experiments on CVA-BUS dataset~\cite{lin2022new} demonstrate that UltraDet, with real-time inference speed, significantly outperforms previous works, reducing about 50\% FPs at a recall rate of 0.90.

Our contributions are four-fold. (1) We identify that the failure of general-purpose detectors on ultrasound videos derives from their incapability of utilizing negative temporal contexts. (2) We propose a novel UltraDet model, incorporating an NTCA module that effectively leverages NTC for FP suppression. (3) We conduct extensive experiments to demonstrate the proposed UltraDet significantly outperforms the previous state-of-the-arts. (4) We release high-quality labels of the CVA-BUS dataset~\cite{lin2022new} to facilitate future research.

\section{Related Works}

\noindent\textbf{Real-Time Video Object Detection} is typically achieved by single-frame detectors, often with temporal information aggregation modules. One-stage detectors~\cite{ge2021yolox,lin2017focal,tian2019fcos,wang2021end} use only intra-frame information, DETR-based detectors~\cite{zhou2022transvod,wang2022ptseformer} and Faster R-CNN-based detectors~\cite{ren2015faster,hu2018relation,zhu2017flow,wu2019sequence,deng2019relation,chen2020memory} are also widely utilized in video object detection. They aggregate temporal information by mining inter-object relationships without considering NTC. 

\noindent\textbf{Ultrasound Lesion Detection}~\cite{liu2019deep} can assist radiologists in clinical practice. Previous works have explored lesion detection in still images~\cite{yap2017automated} and offline videos~\cite{movahedi2020automated,lin2022new,wang2022key}. Real-time video lesion detection is underexplored. In previous works, YOLO series~\cite{wu2021cachetrack,tiyarattanachai2022feasibility} and knowledge distillation~\cite{vaze2020low} are used to speed up inference. However, these works use single-frame detectors or post-process methods while learnable inter-frame aggregation modules are not adopted. Thus their performances are far from satisfactory.

\noindent\textbf{Optical Flow}~\cite{dosovitskiy2015flownet} is used to guide ultrasound segmentation~\cite{nguyen2020end}, motion estimation~\cite{evain2020pilot} and elastography~\cite{peng2018convolution}. For the first time, we use inverse optical flow to guide temporal context information extraction.

\section{Method}

\begin{figure}[htbp]
    \centering
    \includegraphics[width=\textwidth]{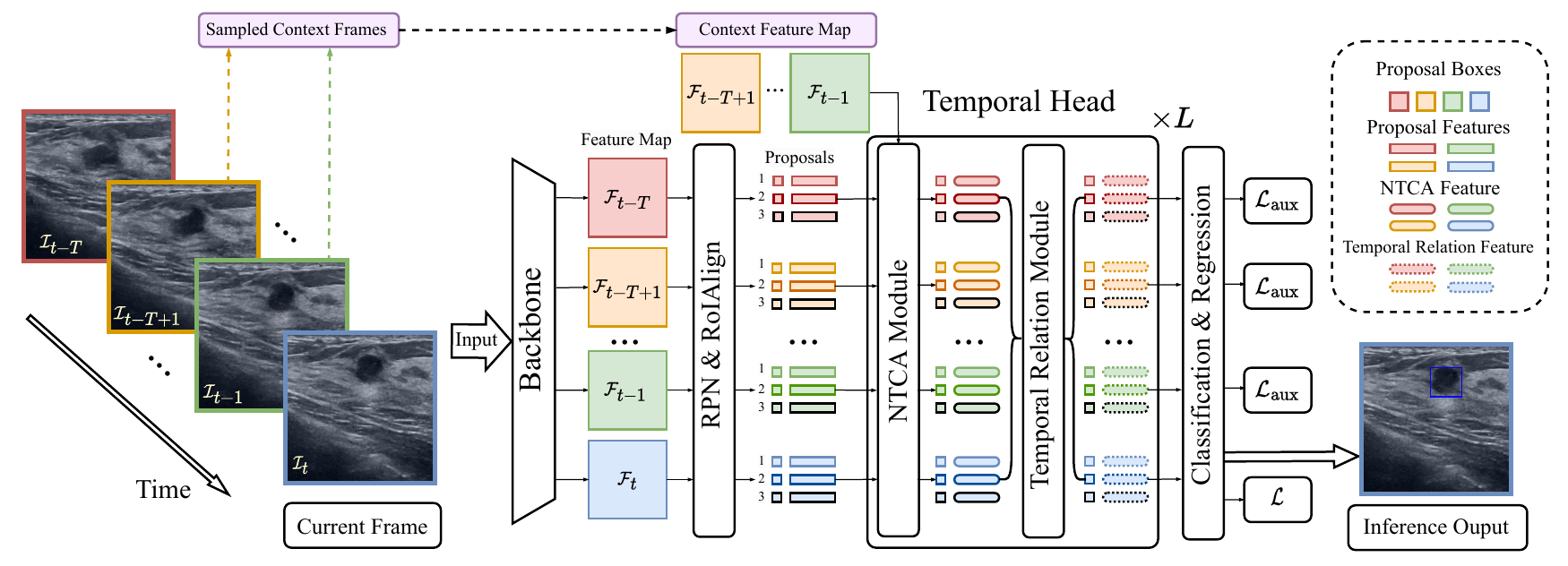}
    \caption{Illustration of UltraDet model. The yellow and green frames are sampled as context frames, and their feature maps are inputs of the NTCA module.}
    \label{fig:model}
\end{figure}

In real-time video lesion detection, given the current frame $\mathcal{I}_t$ and a sequence of $T$ previous frames as $\{\mathcal{I}_{\tau}\}_{\tau=t-T}^{t-1}$, the goal is to detect lesions in $\mathcal{I}_t$ by exploiting the temporal information in previous frames as illustrated in Figure~\ref{fig:model}.

\subsection{Basic Real-Time Detector}\label{subsec:basic_det}

The basic real-time detector comprises three main components: a lightweight backbone (\eg ResNet34~\cite{he2016deep}), a Region Proposal Network (RPN)~\cite{ren2015faster}, and a Temporal Relation head~\cite{deng2019relation}. The backbone is responsible for extracting feature map $\mathcal{F}_{\tau}$ of frame $\mathcal{I}_{\tau}$. The RPN generates proposals consisting of boxes $\mathcal{B}_{\tau}$ and proposal features $\mathcal{Q}_{\tau}$ using RoI Align and average pooling: \begin{equation}
\mathcal{Q}_{\tau}=\operatorname{AvgPool}\left(\operatorname{RoIAlign}(\mathcal{F}_{\tau},\mathcal{B}_{\tau})\right)
\end{equation} where $\tau=t-T,\cdots,t-1,t$. To aggregate temporal information, proposals from all $T+1$ frames are fed into the Temporal Relation head and updated with inter-lesion information extracted via a relation operation~\cite{hu2018relation}: \begin{equation}
\mathcal{Q}^{l}=\mathcal{Q}^{l-1}+\operatorname{Relation}(\mathcal{Q}^{l-1}, \mathcal{B})
\end{equation}
where $l=1,\cdots, L$ represent layer indices, $\mathcal{B}$ and $\mathcal{Q}$ are the concatenation of all $\mathcal{B}_{\tau}$ and $\mathcal{Q}_{\tau}$, and $\mathcal{Q}^{0}=\mathcal{Q}$. We call this basic real-time detector \emph{BasicDet}. The BasicDet is conceptually similar to RDN~\cite{deng2019relation} but does not incorporate relation distillation since the number of lesions and proposals in this study is much smaller than in natural videos.

\subsection{Negative Temporal Context Aggregation}\label{subsec:temp_context} 

\begin{figure}[htbp]
    \centering
    \includegraphics[width=0.8\textwidth]{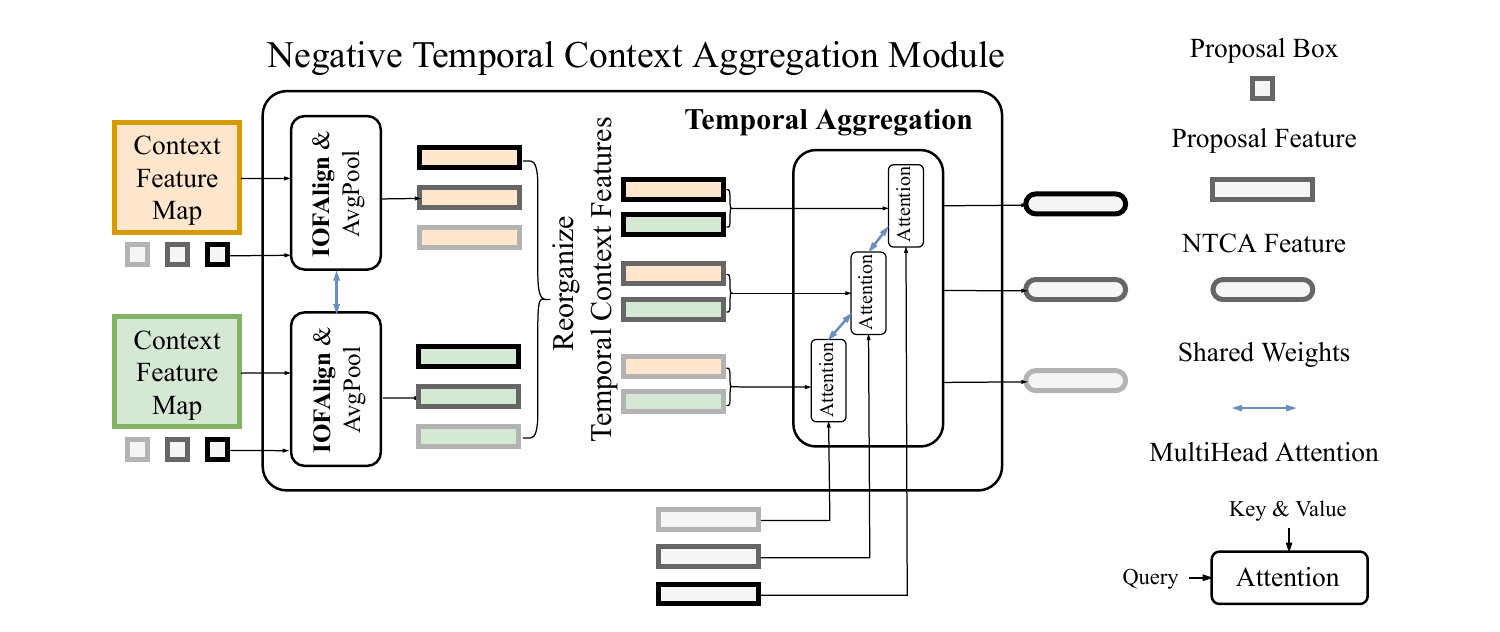}
    \caption{Illustration of the Negative Temporal Context Aggregation module.}
    \label{fig:ntca}
\end{figure}

In this section, we present the Negative Temporal Context Aggregation (NTCA) module. We sample $T_{\text{ctxt}}$ context frames from $T$ previous frames, then extract temporal contexts (TC) from context frames and aggregate them into proposals. We illustrate the NTCA module in Figure~\ref{fig:ntca} and elaborate on details as follows.

\noindent\textbf{Inverse Optical Flow Align} We propose the Inverse Optical Flow Align (IOF Align) to extract TC features. For the current frame $\mathcal{I}_t$ and a sampled context frame $\mathcal{I}_{\tau}$ with $\tau < t$, we extract TC features from the context feature map $\mathcal{F}_{\tau}$ with the corresponding regions. We use inverse optical flow $\mathcal{O}_{t\rightarrow\tau}\in\mathbb{R}^{H\times W\times 2}$ to transform the RoIs from frame $t$ to $\tau$: $\mathcal{O}_{t\rightarrow\tau} = \operatorname{FlowNet}(\mathcal{I}_{t}, \mathcal{I}_{\tau})$ where $H$, $W$ represent height and width of feature maps. The $\operatorname{FlowNet}(\mathcal{I}_{t}, \mathcal{I}_{\tau})$ is a fixed network~\cite{dosovitskiy2015flownet} to predict optical flow from $\mathcal{I}_{t}$ to $\mathcal{I}_{\tau}$. We refer to $\mathcal{O}_{t\rightarrow\tau}$ as \emph{inverse optical flow} because it represents the optical flow in inverse chronological order from $t$ to $\tau$. We conduct IOF Align and average pooling to extract $\mathcal{C}_{t,\tau}$:\begin{equation}
\mathcal{C}_{t,\tau}=\operatorname{AvgPool}\left(\operatorname{IOFAlign}(\mathcal{F}_{\tau}, \mathcal{B}_{t}, \mathcal{O}_{t\rightarrow\tau})\right)
\end{equation} where $\operatorname{IOFAlign}(\mathcal{F}_{\tau}, \mathcal{B}_{t}, \mathcal{O}_{t\rightarrow\tau})$ extracts context features in $\mathcal{F}_\tau$ from deformed grids generated by applying offsets $\mathcal{O}_{t\rightarrow\tau}$ to the original regular grids in $\mathcal{B}_{t}$, which is illustrated in the Figure~\ref{fig:teaser}(b).

\noindent\textbf{Temporal Aggregation} We concatenate $\mathcal{C}_{t, \tau}$ in all $T_{\text{ctxt}}$ context frames to form $\mathcal{C}_{t}$ and enhance proposal features by fusing $\mathcal{C}_{t}$ into $\mathcal{Q}_{t}$:  \begin{equation}
\mathcal{Q}_{\text{ctxt},t}^{l}=\mathcal{Q}_{\text{ctxt},t}^{l-1} + \operatorname{Attention}(\mathcal{Q}_{\text{ctxt},t}^{l-1}, \mathcal{C}_{t}, \mathcal{C}_{t})
\end{equation} where  $l=1,\cdots,L$ represent layer indices, $\mathcal{Q}_{\text{ctxt},t}^{0}=\mathcal{Q}_{t}$, and $\operatorname{Attention}(Q,K,V)$ is Multi-head Attention~\cite{vaswani2017attention}. We refer to the concatenation of all TC-enhanced proposal features in $T+1$ frames as $\mathcal{Q}_{\text{ctxt}}$. To extract consistent TC, the context frames of $T$ previous frames are shared with the current frame.

\subsection{UltraDet For Real-Time Lesion Detection}

We integrate the NTCA module into the BasicDet introduced in Section~\ref{subsec:basic_det} to form the UltraDet model, which is illustrated in Figure~\ref{fig:model}. The head of UltraDet consists of stacked NTCA and relation modules: \begin{equation}
    \mathcal{Q}^l=\mathcal{Q}^{l}_{\text{ctxt}}+\operatorname{Relation}(\mathcal{Q}^{l}_{\text{ctxt}},\mathcal{B}).
\end{equation} During training, we apply regression and classification losses $\mathcal{L}=\mathcal{L}_{\text{reg}} + \mathcal{L}_{\text{cls}}$ to the current frame. To improve training efficiency, we apply auxiliary losses $\mathcal{L}_{\text{aux}}=\mathcal{L}$ to all previous $T$ frames. During inference, the UltraDet model uses the current frame and $T$ previous frames as inputs and generates predictions only for the current frame. This design endows the UltraDet with the ability to perform real-time lesion detection.

\section{Experiments}

\subsection{Dateset}

\noindent\textbf{CVA-BUS Dateset} We use the open source CVA-BUS dataset that consists of 186 valid videos, which is proposed in CVA-Net~\cite{lin2022new}. We split the dataset into train-val (154 videos) and test (32 videos) sets. In the train-val split, there are 21423 frames with 170 lesions. In the test split, there are 3849 frames with 32 lesions. We focus on the lesion detection task and do not utilize the benign/malignant classification labels provided in the original dataset.

\noindent\textbf{High-quality Labels} The bounding box labels provided in the original CVA-BUS dataset are unsteady and sometimes inaccurate, leading to jiggling and inaccurate model predictions. We provide a new version of high-quality labels that are re-annotated by experienced radiologists. We reproduce all baselines using our high-quality labels to ensure a fair comparison. Visual comparisons of two versions of labels are available in supplementary materials. To facilitate future research, we will release these high-quality labels.

\begin{table}[htbp]
\centering
\caption{Quantitative results of real-time lesion detection on CVA-BUS~\cite{lin2022new}.}
\begin{tabular}{lc|cccccccc}
\hline
Model & Type & Pr\scriptsize{80} & Pr\scriptsize{90} & FP\scriptsize{80} & FP\scriptsize{90} & AP\scriptsize{50} & R@16 & FPS \\
\hline\hline
& & \multicolumn{8}{c}{One-Stage Detectors}\\
\hline
YOLOX~\cite{ge2021yolox} & Image & 69.7\tiny{3.7} & 43.4\tiny{7.7} & 23.8\tiny{4.8} & 87.6\tiny{24.5} & 80.4\tiny{1.6} & 97.5\tiny{0.5} & \textbf{59.8} \\
RetinaNet~\cite{lin2017focal} & Image & 75.7\tiny{2.5} & 57.2\tiny{2.9} & 9.3\tiny{2.0} & 32.8\tiny{6.5} & 84.5\tiny{1.0} & 95.1\tiny{0.6} & 53.6 \\
FCOS~\cite{tian2019fcos} & Image & 87.2\tiny{2.2} & 72.2\tiny{5.1} & 11.0\tiny{2.4} & 23.0\tiny{3.7} & 89.5\tiny{1.4} & 98.8\tiny{0.3} & 56.1 \\
DeFCN~\cite{wang2021end} & Image & 81.5\tiny{1.8} & 67.5\tiny{2.3} & 21.1\tiny{3.2} & 33.4\tiny{4.3} & 86.4\tiny{1.3} & \textbf{99.3}\tiny{0.3} & 51.2 \\
Track-YOLO~\cite{wu2021cachetrack} & Video & 75.1\tiny{2.7} & 47.0\tiny{3.1} & 18.1\tiny{1.9} & 74.2\tiny{14.7} & 80.1\tiny{1.0} & 94.7\tiny{0.9} & 46.0\\
\hline
& & \multicolumn{8}{c}{DETR-Based Detectors}\\
\hline
DeformDETR~\cite{zhu2020deformable} & Image & 90.1\tiny{3.2} & 72.7\tiny{10.6} & 5.6\tiny{2.2} & 37.8\tiny{20.9} & 90.5\tiny{2.0} & 98.7\tiny{0.3} & 33.8 \\
TransVOD~\cite{zhou2022transvod} & Video & 92.5\tiny{2.2} & 77.5\tiny{7.2} & 3.1\tiny{1.3} & 23.7\tiny{11.5} & 90.1\tiny{1.8} & 98.4\tiny{0.4} & 24.2 \\
CVA-Net~\cite{lin2022new} & Video & 92.3\tiny{2.6} & 80.2\tiny{6.1} & 4.7\tiny{2.6} & 19.6\tiny{5.6} & \textbf{91.6}\tiny{1.9} & 98.6\tiny{0.8} & 23.1 \\
PTSEFormer~\cite{wang2022ptseformer} & Video & 93.3\tiny{1.9} & 85.4\tiny{6.0} & 2.8\tiny{1.1} & 12.5\tiny{9.8} & 91.5\tiny{1.6} & 97.9\tiny{1.2} & 9.1 \\
\hline
& & \multicolumn{8}{c}{FasterRCNN-Based Detectors}\\
\hline
FasterRCNN~\cite{ren2015faster} & Image & 91.3\tiny{0.9} & 75.2\tiny{3.6} & 6.9\tiny{1.4} & 34.4\tiny{6.7} & 88.0\tiny{1.4} & 92.4\tiny{1.0} & 49.2 \\
RelationNet~\cite{hu2018relation} & Image & 91.4\tiny{1.3} & 79.2\tiny{2.9} & 6.2\tiny{2.0} & 24.4\tiny{5.6} & 87.6\tiny{1.7} & 92.4\tiny{0.9} & 42.7 \\
FGFA~\cite{zhu2017flow} & Video & 92.9\tiny{1.5} & 82.2\tiny{4.1} & 4.4\tiny{1.6} & 13.3\tiny{3.7} & 90.5\tiny{1.1} & 93.6\tiny{0.9} & 33.8\\
SELSA~\cite{wu2019sequence} & Video & 91.6\tiny{1.7} & 80.2\tiny{2.5} & 7.5\tiny{1.5} & 23.3\tiny{5.5} & 89.2\tiny{1.1} & 92.6\tiny{0.8} & 43.8 \\
MEGA~\cite{chen2020memory} & Video & 93.9\tiny{1.5} & 86.9\tiny{2.3} & 3.1\tiny{1.7} & 11.7\tiny{3.0} & 90.9\tiny{1.0} & 93.6\tiny{0.7} & 40.2 & \\
BasicDet (RDN)~\cite{deng2019relation} & Video & 92.4\tiny{1.0} & 83.6\tiny{2.2} & 3.8\tiny{1.2} & 13.4\tiny{3.2} & 88.7\tiny{1.4} & 92.7\tiny{0.6} & 42.2 \\
UltraDet (Ours) & Video & \textbf{95.7}\tiny{1.2} & \textbf{90.8}\tiny{1.4} & \textbf{1.9}\tiny{0.4} & \textbf{5.7}\tiny{1.6} & \textbf{91.6}\tiny{1.6} & 93.8\tiny{1.3} & 30.4 \\
\hline
\end{tabular}
\label{tab:sota_comparison}
\end{table}

\subsection{Evaluation Metrics}

\noindent\textbf{{Pr\scriptsize{80}}, {Pr\scriptsize{90}}} In clinical applications, it is important for detection models to be sensitive. So we provide frame-level precision values with high recall rates of 0.80 and 0.90, which we denote as {Pr\scriptsize{80}} and {Pr\scriptsize{90}}, respectively. 

\noindent\textbf{{FP\scriptsize{80}}, {FP\scriptsize{90}}} We further report lesion-level FP rates as critical metrics. Frame-level FPs are linked by IoU scores to form FP sequences~\cite{wu2021cachetrack}. The number of FP sequences per minute at recall rates of 0.80 and 0.90 are reported as {FP\scriptsize{80}} and {FP\scriptsize{90}}, respectively. The unit of lesion-level FP rates is seq/min.

\noindent\textbf{{AP\scriptsize{50}}} We provide {AP\scriptsize{50}} instead of mAP or {AP\scriptsize{75}} because the IoU threshold of 0.50 is sufficient for lesion localization in clinical practice. Higher thresholds like 0.75 or 0.90 are impractical due to the presence of blurred lesion edges.

\noindent\textbf{R@16} To evaluate the highest achievable sensitivity, we report the frame-level average recall rates of Top-16 proposals, denoted as R@16.

\subsection{Implementation Details}\label{subsec:impl_detail}

\noindent\textbf{UltraDet Settings} We use FlowNetS~\cite{dosovitskiy2015flownet} as the fixed FlowNet in IOF Align and share the same finding with previous works~\cite{peng2018convolution,evain2020pilot,nguyen2020end} that the FlowNet trained on natural datasets generalizes well on ultrasound datasets. We set the pooling stride in the FlowNet to 4, the number of UltraDet head layers $L=2$, the number of previous frames $T=15$ and $T_{\text{ctxt}}=2$, and the number of proposals is $16$. We cached intermediate results of previous frames and reuse them to speed up inference. Other hyper-parameters are listed in supplementary materials.

\noindent\textbf{Shared Settings} All models are built in PyTorch framework and trained using eight NVIDIA GeForce RTX 3090 GPUs. We use ResNet34~\cite{he2016deep} as backbones and set the number of training iterations to 10,000. We set the feature dimensions of detection heads to 256 and baselines are re-implemented to utilize only previous frames. We refer to our code for more details.

\subsection{Main Results}\label{subsec:main_result}

\noindent\textbf{Quantitative Results} We compare performances of real-time detectors with the UltraDet in Table~\ref{tab:sota_comparison}. We perform 4-fold cross-validation and report the mean values and standard errors on the test set to mitigate fluctuations. The UltraDet outperforms all previous state-of-the-art in terms of precision and FP rates. Especially, the {Pr\scriptsize{90}} of UltraDet achieves 90.8\%, representing a 5.4\% absolute improvement over the best competitor, PTSEFormer~\cite{wang2022ptseformer}. Moreover, the {FP\scriptsize{90}} of UltraDet is 5.7 seq/min, reducing about 50\% FPs of the best competitor, PTSEFormer. Although CVA-Net~\cite{lin2022new} achieve comparable {AP\scriptsize{50}} with our method, we significantly improve precision and FP rates over the CVA-Net~\cite{lin2022new}.

\begin{figure}[htbp]
    \centering
    \includegraphics[width=\textwidth]{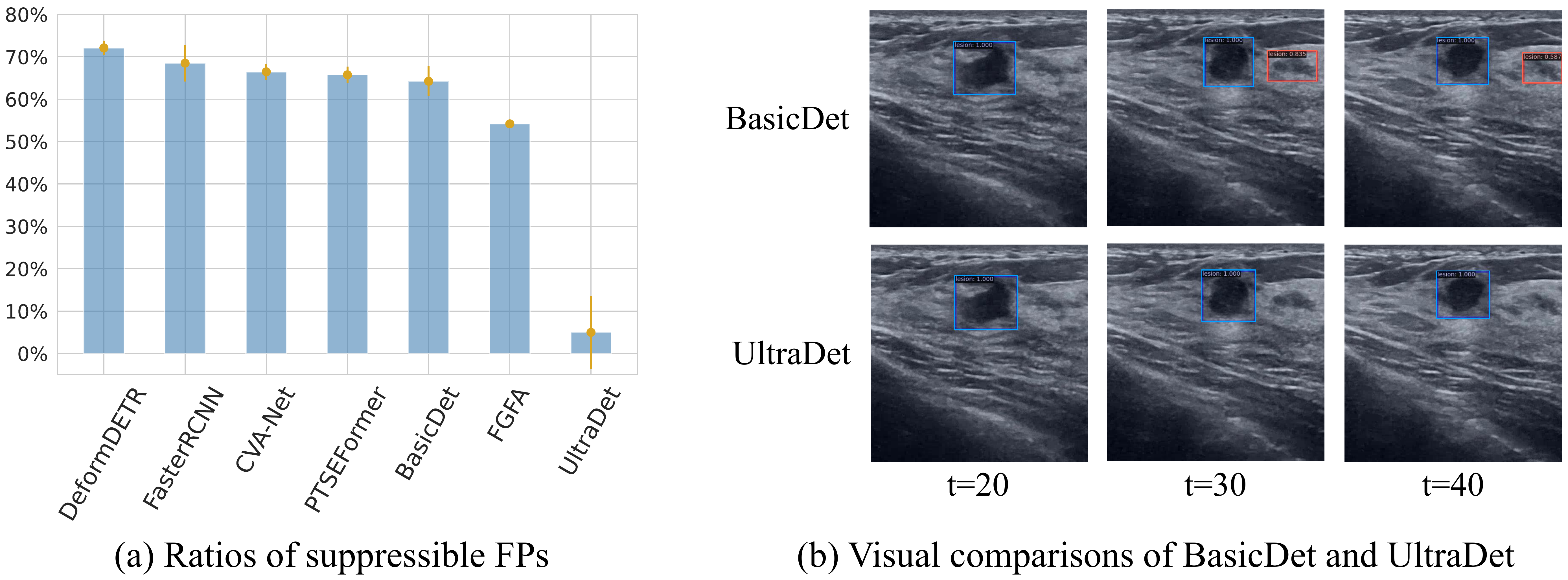}
    \caption{(a) Ratios of FPs that are suppressible by leveraging NTC. (b) Visual comparisons of BasicDet and UltraDet prediction results at recall 0.90. Blue boxes are true positives and red boxes are FPs.}
    \label{fig:visualization}
\end{figure}

\noindent\textbf{Importance of NTC} In Figure~\ref{fig:visualization}(a), we illustrate the FP ratios that can be suppressed by using NTC. The determination of whether FPs can be inhibited by NTC is based on manual judgments of experienced radiologists. We find that about 50\%$\sim$70\% FPs of previous methods are suppressible. However, by utilizing NTC in our UltraDet, we are able to effectively prevent this type of FPs.

\noindent\textbf{Inference Speed} We run inference using one NVIDIA GeForce RTX 3090 GPU and report the inference speed in Table~\ref{tab:sota_comparison}. The UltraDet achieves an inference speed of 30.4 FPS and already meets the 30 FPS requirement. Using TensorRT, we further optimize the speed to 35.2 FPS, which is sufficient for clinical applications~\cite{vaze2020low}.

\noindent\textbf{Qualitative Results} Figure~\ref{fig:visualization}(b) visually compares BasicDet and UltraDet. The BasicDet reports FPs at $t=30$ and $40$ as it fails to leverage NTC when $t=20$, while the UltraDet successfully suppresses FPs with the NTCA module.

\subsection{Ablation Study}

\begin{table}[htbp]
\centering
\caption{Ablation study of each NTCA sub-module.}
\begin{tabular}{cc|cccccccccc}
\hline
IOFAlign & TempAgg & Pr\scriptsize{80} & Pr\scriptsize{90} & FP\scriptsize{80} & FP\scriptsize{90} & AP\scriptsize{50} & R@16 & FPS \\
\hline\hline
- & - & 92.4\tiny{1.0} & 83.6\tiny{2.2} & 3.8\tiny{1.2} & 13.4\tiny{3.2} & 88.7\tiny{1.4} & 92.7\tiny{0.6} & 42.2 \\
- & $\checkmark$ & 93.7\tiny{1.8} & 84.3\tiny{1.4} & 3.4\tiny{1.0} & 12.5\tiny{0.8} & 90.0\tiny{1.9} & 93.0\tiny{1.3} & 37.2\\
$\checkmark$ & - & 94.5\tiny{2.3} & 88.7\tiny{2.2} & 2.6\tiny{0.6} & 9.0\tiny{1.5} & 90.5\tiny{1.9} & 92.9\tiny{1.4} & 32.3\\
$\checkmark$ & $\checkmark$ & \textbf{95.7}\tiny{1.2} & \textbf{90.8}\tiny{1.4} & \textbf{1.9}\tiny{0.4} & \textbf{5.7}\tiny{1.6} & \textbf{91.6}\tiny{1.6} & \textbf{93.8}\tiny{1.3} & 30.4\\
\hline
\end{tabular}
\label{tab:ablate}
\end{table}

\noindent\textbf{Effectiveness of each sub-module} We ablate the effectiveness of each sub-module of the NTCA module in Table~\ref{tab:ablate}. Specifically, we replace the IOF Align with an RoI Align and the Temporal Aggregation with a simple average pooling in the temporal dimension. The results demonstrate that both IOF Align and Temporal Aggregation are crucial, as removing either of them leads to a noticeable drop in performance.

\begin{table}[htbp]
\centering
\caption{Design of the NTCA Module.}
\begin{tabular}{l|ccccccccccc}
\hline
Num & Pr\scriptsize{80} & Pr\scriptsize{90} & FP\scriptsize{80} & FP\scriptsize{90} & AP\scriptsize{50} & R@16 & FPS \\
\hline\hline
Feature-level & 94.0\tiny{0.9} & 84.6\tiny{2.9} & 2.9\tiny{0.8} & 11.7\tiny{3.0} & 90.8\tiny{0.7} & 93.3\tiny{0.6} & 30.6\\
RoI-level & \textbf{95.7}\tiny{1.2} & \textbf{90.8}\tiny{1.4} & \textbf{1.9}\tiny{0.4} & \textbf{5.7}\tiny{1.6} & \textbf{91.6}\tiny{1.6} & \textbf{93.8}\tiny{1.3} & 30.4\\
Both-level & 94.6\tiny{1.0} & 88.7\tiny{1.8} & 2.5\tiny{0.9} & 7.9\tiny{2.4} & 90.8\tiny{1.5} & \textbf{93.8}\tiny{0.9} & 26.9\\
\hline
\end{tabular}
\label{tab:design}
\end{table}

\noindent\textbf{Design of the NTCA module} Besides RoI-level TC aggregation in UltraDet, feature-level aggregation is also feasible. We plug the optical flow feature warping proposed in FGFA~\cite{zhu2017flow} into the BasicDet and report the results in Table~\ref{tab:design}. We find RoI-level aggregation is more effective than feature-level, and both-level aggregation provides no performance gains. This conclusion agrees with radiologists' skills to focus more on local regions instead of global information.

\section{Conclusion}

In this paper, we address the clinical challenge of real-time ultrasound lesion detection. We propose a novel Negative Temporal Context Aggregation (NTCA) module, imitating radiologists' diagnosis processes to suppress FPs. The NTCA module leverages negative temporal contexts that are essential for FP suppression but ignored in previous works, thereby being more effective in suppressing FPs. We plug the NTCA module into a BasicDet to form the UltraDet model, which significantly improves the precision and FP rates over previous state-of-the-arts while achieving real-time inference speed. The UltraDet has the potential to become a real-time lesion detection application and assist radiologists in more accurate cancer diagnosis in clinical practice.

\paragraph{\emph{\textbf{Acknowledgements.}}} This work is supported by National Key R\&D Program of China (2022ZD0114900) and National Science Foundation of China (NSFC62276005).

%
%
%
\bibliographystyle{splncs04}
\bibliography{mybibliography}
\newpage
\title{Supplementary Material}
\titlerunning{Supplementary Material}
\author{}
\institute{}
\authorrunning{H. Yu et al.}

\maketitle

\begin{figure}[H]
    \centering
    \includegraphics[width=\textwidth]{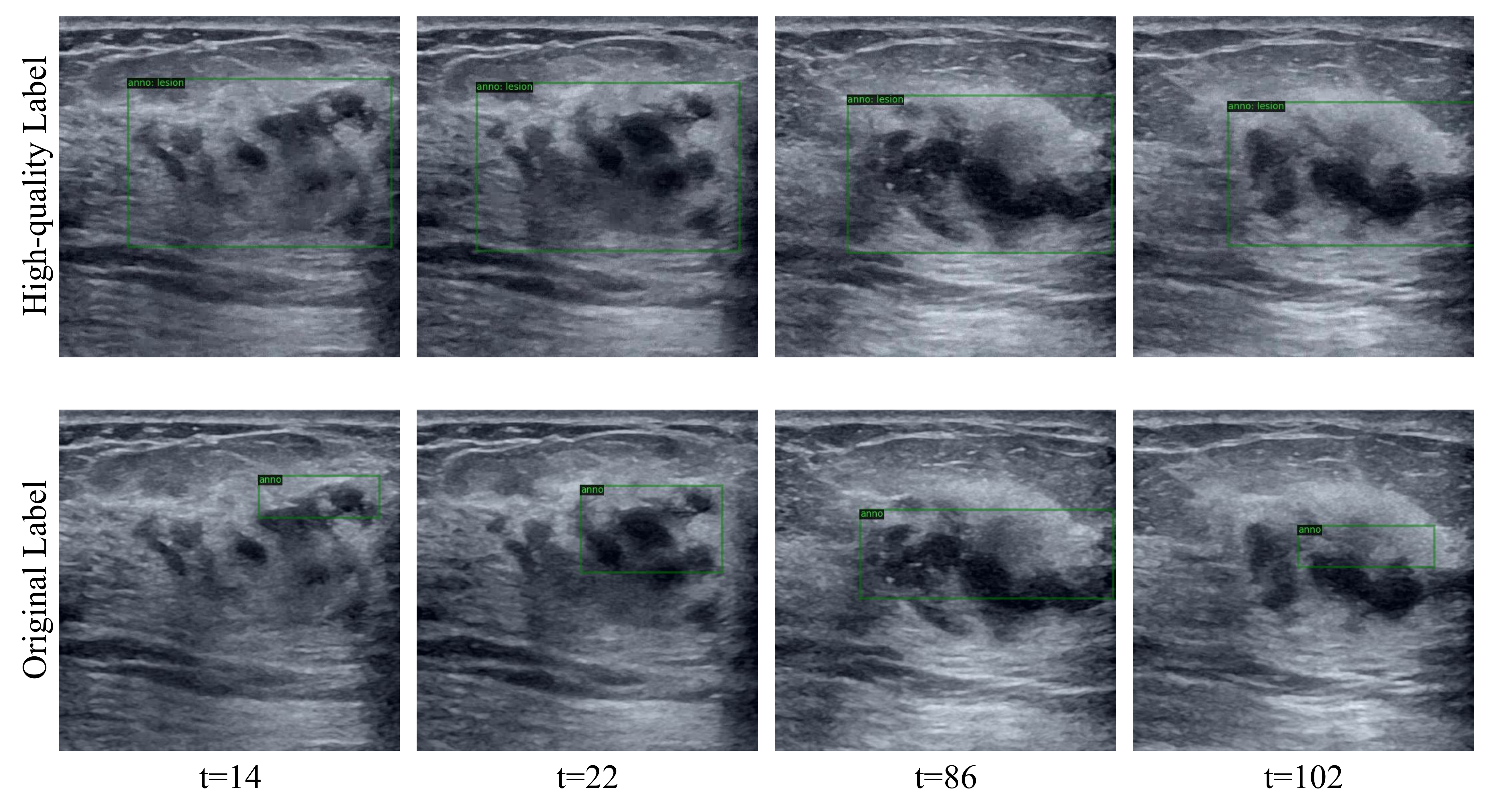}
    \caption{Visual comparison of high-quality labels and original labels. The high-quality labels are more steady and more accurate. We provide the high-quality labels in \textbf{bus\_data\_cva\_new} folder in ".json" format.}
    \label{fig:label}
    \vspace{-30pt}
\end{figure}

\begin{figure}[H]
    \centering
    \includegraphics[width=\textwidth]{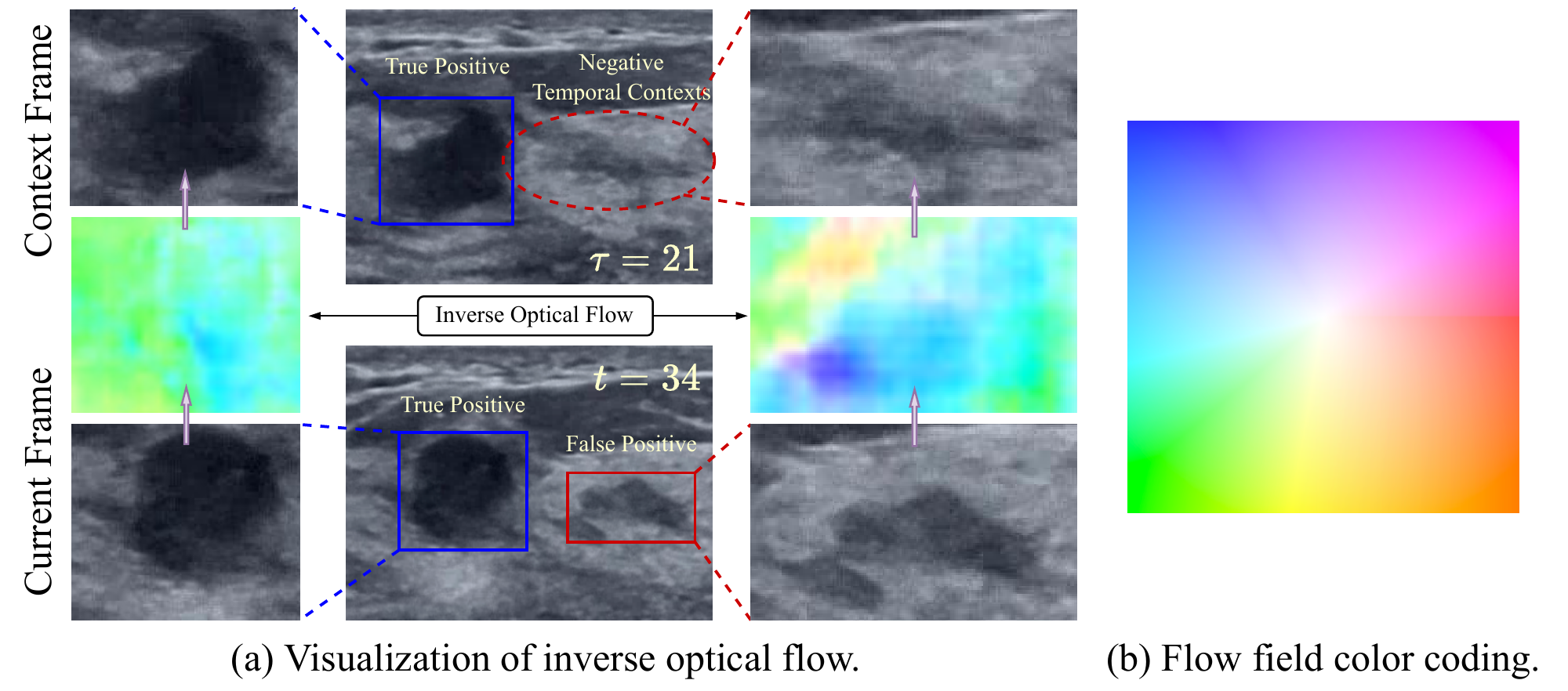}
    \caption{(a) Visualization of inverse optical flow $\mathcal{O}_{t\rightarrow\tau}=\operatorname{FlowNet}(\mathcal{I}_{t}, \mathcal{I}_{\tau})$ applied in the IOF Align. (b) Flow field color used in (a). The color of each pixel represents the movement from the central pixel to it.}
    \label{fig:optical_flow}
    \vspace{-30pt}
\end{figure}

\begin{table}[H]
\centering
\caption{Hyper-parameters used in the main paper. We resize input images to 900,000 pixels and apply random scale during training. The source codes of the UltraDet and all baselines are attached in the supplementary materials.} 
    \begin{tabular}{lllllll}
    \hline
    Learning Rate (LR) & Optimizer & $\beta_1$ & $\beta_2$ & Weight Decay\\ 
    \hline
    2e-4 & AdamW & 0.9 & 0.99 & 1e-4 \\
    \hline\hline
    LR schedule & Decay Steps & Input Resize & Training Scale & Test Scale \\
    \hline
    Step & (6,000, 8,000) &  900,000 & (0.3,1.5) & 1.0\\
    \hline
    \end{tabular}
    \label{tab:hyperparameter}
    \vspace{-10pt}
\end{table}

\begin{table}[H]
\centering
\caption{Quantitative results of real-time lesion detection. ResNet18 is used in all models as the backbone. The performance gains of the UltraDet are still significant with the smaller ResNet18 backbone.}
\begin{tabular}{lc|ccccccccccc}
\hline
Model & Type &  Pr\scriptsize{80} & Pr\scriptsize{90} & FP\scriptsize{80} & FP\scriptsize{90} & AP\scriptsize{50} & R@16 & FPS \\
\hline\hline
& & \multicolumn{7}{c}{One-Stage Detector}\\
\hline
YOLOX & Image & - & - & - & - & 43.5\tiny{3.8} & 84.9\tiny{3.4} & 96.0\\
RetinaNet & Image & 57.2\tiny{4.4} & 31.7\tiny{4.4} & 27.7\tiny{5.6} & 79.5\tiny{16.1} & 76.8\tiny{1.9} & 94.7\tiny{0.8} & 85.7\\
FCOS & Image & 66.2\tiny{3.9} & 45.6\tiny{4.3} & 29.8\tiny{5.2} & 53.9\tiny{11.5} & 80.1\tiny{2.1} & \textbf{98.9}\tiny{0.2} & 86.2\\
DeFCN & Image & 56.6\tiny{3.2} & 40.0\tiny{4.7} & 45.5\tiny{5.8} & 70.4\tiny{12.3} & 73.8\tiny{2.4} & \textbf{98.9}\tiny{0.2} & 79.1\\
\hline
& & \multicolumn{7}{c}{DETR-Based Detectors}\\
\hline
DeformDETR & Image & 88.2\tiny{2.9} & 67.7\tiny{6.8} & 7.3\tiny{2.6} & 47.0\tiny{18.0} & 88.7\tiny{2.3} & 98.5\tiny{0.5} & 40.3 \\
CVA-Net & Video & 88.4\tiny{2.2} & 68.3\tiny{7.8} & 6.6\tiny{2.0} & 42.1\tiny{17.5} & 89.2\tiny{2.1} & 98.7\tiny{0.8} & 32.2 \\
TransVOD & Video & 91.0\tiny{4.1} & 75.4\tiny{13.9} & 5.3\tiny{3.4} & 35.6\tiny{31.5} & 90.4\tiny{2.9} & 98.5\tiny{0.3} & 31.5\\
PTSEFormer & Video & 91.8\tiny{1.7} & 80.8\tiny{5.0} & 3.5\tiny{0.8} & 18.0\tiny{8.1} & \textbf{91.3}\tiny{1.3} & 98.3\tiny{0.6} & 10.2\\
\hline
& & \multicolumn{7}{c}{FasterRCNN-Based Detectors}\\
\hline
FasterRCNN & Image & 89.0\tiny{1.7} & 76.9\tiny{2.7} & 8.8\tiny{1.4} & 27.8\tiny{4.7} & 89.3\tiny{1.3} & 95.5\tiny{0.4} & 68.0 \\
RelationNet & Image & 92.4\tiny{1.6} & 80.4\tiny{3.1} & 6.3\tiny{1.5} & 25.2\tiny{5.7} & 89.1\tiny{1.4} & 92.6\tiny{1.0} & 58.8\\
FGFA & Video & 88.2\tiny{1.6} & 75.7\tiny{3.8} & 6.3\tiny{1.6} & 18.2\tiny{5.0} & 89.0\tiny{1.4} & 95.6\tiny{0.6} & 43.9\\
SELSA & Video & 90.6\tiny{1.9} & 80.5\tiny{2.2} & 8.5\tiny{1.8} & 21.3\tiny{3.8} & 89.8\tiny{1.7} & 95.2\tiny{0.7} & 62.2\\
MEGA & Video & 92.8\tiny{2.1} & 85.0\tiny{4.0} & 3.7\tiny{1.1} & 11.1\tiny{4.5} & 90.6\tiny{2.3} & 95.4\tiny{0.9} & 56.3\\
BasicDet (RDN) & Video & 92.5\tiny{1.3} & 84.2\tiny{1.9} & 3.8\tiny{1.4} & 10.7\tiny{2.8} & 90.2\tiny{1.3} & 95.1\tiny{0.4} & 56.8\\
UltraDet & Video & \textbf{93.8}\tiny{1.6} & \textbf{87.6}\tiny{3.4} & \textbf{3.0}\tiny{1.6} & \textbf{7.6}\tiny{3.0} & 90.5\tiny{2.1} & 95.3\tiny{0.6} & 37.3\\
\hline
\end{tabular}
\vspace{-10pt}
\end{table}

\begin{table}[H]
\centering
\caption{Ablation Study. We ablate the number of context frames $T_{\text{ctxt}}$ sampled from previous frames and find that performances of the UltraDet reachs the best when $T_{\text{ctxt}}=2$. Redundant context frames may hurt the UltraDet performances.}
\begin{tabular}{c|cccccccc}
\hline
        Num & Pr\scriptsize{80} & Pr\scriptsize{90} & FP\scriptsize{80} & FP\scriptsize{90} & AP\scriptsize{50} & R@16 \\
        \hline\hline
        1 & 94.2\tiny{1.9} & 88.5\tiny{1.9} & 2.3\tiny{0.8} & 7.8\tiny{1.2} & 90.5\tiny{2.0} & 93.3\tiny{1.5} \\
        2 & \textbf{95.7}\tiny{1.2} & \textbf{90.8}\tiny{1.4} & \textbf{1.9}\tiny{0.4} & \textbf{5.7}\tiny{1.6} & \textbf{91.6}\tiny{1.6} & \textbf{93.8}\tiny{1.3}\\
        4 & 95.4\tiny{1.7} & 90.1\tiny{2.9} & 2.6\tiny{1.4} & 6.9\tiny{2.1} & 91.4\tiny{1.7} & 93.7\tiny{1.0} \\
        \hline
\end{tabular}
\end{table}

\begin{table}[H]
\centering
\caption{Ablation Study. We compare the Relation module and Attention module. Attention module outperforms Relation module in frame-level metrics but performs worse in lesion-level metrics.}
\begin{tabular}{c|cccccccc}
\hline
        Num & Pr\scriptsize{80} & Pr\scriptsize{90} & FP\scriptsize{80} & FP\scriptsize{90} & AP\scriptsize{50} & R@16  \\
        \hline\hline
        Attention & \textbf{96.1}\tiny{0.4} & \textbf{91.4}\tiny{1.4} & 2.3\tiny{1.3} & 6.4\tiny{2.4} & \textbf{92.6}\tiny{0.5} & \textbf{94.1}\tiny{0.5}\\
        Relation & 95.7\tiny{1.2} & 90.8\tiny{1.4} & \textbf{1.9}\tiny{0.4} & \textbf{5.7}\tiny{1.6} & 91.6\tiny{1.6} & 93.8\tiny{1.3} \\
        \hline
\end{tabular}
\end{table}

\begin{table}[H]
\centering
\caption{Ablation Study. We ablate the number of NTCA modules. }
\begin{tabular}{c|cccccccc}
\hline
        Num & Pr\scriptsize{80} & Pr\scriptsize{90} & FP\scriptsize{80} & FP\scriptsize{90} & AP\scriptsize{50} & R@16  \\
        \hline\hline
        0 & 92.4\tiny{1.0} & 83.6\tiny{2.2} & 3.8\tiny{1.2} & 13.4\tiny{3.2} & 88.7\tiny{1.4} & 92.7\tiny{0.6}\\
        1 & 94.0\tiny{1.1} & 88.4\tiny{2.5} & 2.3\tiny{0.8} & 7.2\tiny{2.8} & 90.5\tiny{2.1} & 93.3\tiny{1.2}\\
        2 & \textbf{95.7}\tiny{1.2} & \textbf{90.8}\tiny{1.4} & \textbf{1.9}\tiny{0.4} & \textbf{5.7}\tiny{1.6} & \textbf{91.6}\tiny{1.6} & \textbf{93.8}\tiny{1.3} \\
        \hline
\end{tabular}
\end{table}

\begin{table}[H]
\centering
\caption{Ablation Study. We ablate the effectiveness of auxiliary losses. The performance will drop dramatically if we do not use auxiliary losses.}
\begin{tabular}{c|cccccccc}
\hline
        Aux-Loss & Pr\scriptsize{80} & Pr\scriptsize{90} & FP\scriptsize{80} & FP\scriptsize{90} & AP\scriptsize{50} & R@16  \\
        \hline\hline
        $\times$ & 90.7\tiny{7.5} & 76.6\tiny{14.4} & 5.1\tiny{3.3} & 24.5\tiny{17.8} & 87.9\tiny{4.7} & 91.7\tiny{1.3}\\
        $\checkmark$ & \textbf{95.7}\tiny{1.2} & \textbf{90.8}\tiny{1.4} & \textbf{1.9}\tiny{0.4} & \textbf{5.7}\tiny{1.6} & \textbf{91.6}\tiny{1.6} & \textbf{93.8}\tiny{1.3}\\
        \hline
\end{tabular}
\end{table}


\end{document}